\documentclass{article}
\sloppypar

\usepackage{times}
\usepackage[T1]{fontenc} 
\usepackage[latin1]{inputenc}
\usepackage[english]{babel}
\usepackage{exscale}
\usepackage[centertags]{amsmath}
\usepackage{graphicx}
\usepackage{epstopdf}

\usepackage{subcaption}
\usepackage{amssymb}
\usepackage{longtable}
\usepackage{enumitem}
\usepackage{wasysym}

\newlength\figureheight
\newlength\figurewidth

\begin{document}
\thispagestyle{empty}

\renewcommand{\tablename}{\small{Tabelle}}
\renewcommand{\figurename}{\small{Bild}}

\begin{center}
{
\Large \textbf{Datenqualität in Regressionsproblemen}}
\\[3ex]
{\large \textbf{Technischer Bericht, Version 1.0}}
\\[3ex]
{\large \textbf{Wolfgang Doneit$^{1}$, Ralf Mikut$^{1}$, Markus Reischl$^{1}$}}
\\[1ex]
$^{1}$Karlsruher Institut für Technologie, Institut für Angewandte Informatik\\
                       E-Mail: wolfgang.doneit@kit.edu, ralf.mikut@kit.edu, markus.reischl@kit.edu\\
\end{center}
\section{Motivation}
Datenbasierte Modelle mit reellwertigem Ausgang werden als Stellvertreter-Modelle in Optimierungsproblemen, für modellprädiktive Regelungen, Zeitreihenprognosen u.v.m. verwendet. Wir bezeichnen solche Modelle als Regressionen bzw. Regressionsmodelle\footnote{auf die Begriffsdefinition der Regression wird in diesem Beitrag nicht weiter eingegangen, der Begriff dient zunächst lediglich der Abgrenzung zu Klassifikatoren mit nominalskaliertem Ausgang.}. Regressionsmodelle werden z.B. mit Hilfe der aus der Statistik bekannten multilinearen Regression oder Künstlichen Neuronalen Netzen erstellt. 
Zur Modellbildung stehen sogenannte Datentupel als Lerndaten zur Verfügung, die jeweils einem Vektor mit Eingangsdaten einen skalaren Wert der Zielgröße zuordnen. Ziel der Regression ist die Abbildung des funktionalen Zusammenhangs zwischen Eingangsgrößen und Zielgröße.

In der Modellbildung wird eine geeignete Modellstruktur, bzw. Modell\-komplexität gesucht, und ihre freien Parameter werden an die Daten angepasst. In den meisten Fällen wird dazu die Methode der kleinsten Fehlerquadrate verwendet. Außerdem gibt es Erweiterungen der Methode der kleinsten Fehlerquadrate, um verschiedenen Einschränkungen in der Datenqualität gerecht zu werden. Beispiele für bekannte Einschränkungen in der Datenqualität in Regressionsproblemen sind Ausreißer~\cite{Rousseeuw05}, Heteroskedastizität~\cite{Nealen04,Koenker82}, Kollinearität~\cite{Geladi86,Jolliffe02} und fehlerbehaftete Eingangsgrößen~\cite{VanHuffel91}.

Regressionen werden zunehmend auf Datensätzen angewendet, deren Eingangsvektoren nicht durch eine statistische Versuchsplanung~\cite{Bandemer77} festgelegt wurden\footnote{Die Versuchsplanung stellt eine gleichmäßige Verteilung der Eingangsdaten sicher, um alle Zustände eines betrachteten Systems zu erfassen.}. Stattdessen werden die Daten beispielsweise durch die passive Beobachtung technischer Systeme gesammelt. Damit bilden bereits die Eingangsdaten Phänomene des Systems ab und widersprechen statistischen Verteilungsannahmen. Die Verteilung der Eingangsdaten hat Einfluss auf die Zuverlässigkeit eines Regressionsmodells. Wir stellen deshalb Bewertungskriterien für einige typische Phänomene in Eingangsdaten von Regressionen vor und zeigen ihre Funktionalität anhand simulierter Benchmarkdatensätze.

\section{Methoden}
\subsection{Allgemeines}
In den folgenden Abschnitten werden Bewertungskriterien vorgestellt, die sich ausschließlich auf die uni- und bivariaten Verteilungen der Eingangsdaten beziehen und nicht die Zielgröße berücksichtigen. Sie quantifizieren verschiedene Phänomene in den Eingangsdaten und sind daher als Ergänzung zur herkömmlichen Merkmalsbewertung für Regressionsmodelle zu verstehen. Auf multivariate Verfahren wird aufgrund des Fluchs der Dimensionalität und zu Gunsten der Interpretierbarkeit der Kriterien verzichtet. Die quantifizierten  Einschränkungen der Datenqualität sind damit in Histogrammen und Streuwolkendiagrammen zu erkennen. Anders als die in~\cite{Wilkinson05} vorgestellten sogenannten "`Scagnostics"' werden in den hier vorgestellten Bewertungskriterien keine Maße aus der Graphentheorie verwendet. Außerdem liegt der Schwerpunkt auf Phänomenen, die im Kontext der Regressionen nützlich und interpretierbar sind.

\subsection{Begriffe und Symbole}
Stehen $N$ Datentupel als Lerndaten zur Verfügung, die jeweils einem $p$-dimensionalen Eingangsvektor (mit den Ausprägungen für die Eingangsgrößen $x_1,\hdots,x_p$) einen skalaren Wert der Zielgröße $y$ zuordnen, dann sind die Eingangsdaten gegeben als Datenmatrix $\boldsymbol{X}^{N\times p}$, wobei jede Zeile einem Eingangsvektor $\boldsymbol{x}_i^T, i=1,\hdots,N$ entspricht.

Die Bewertungskriterien liegen zur besseren Interpretierbarkeit im Intervall $[0,1]$. Ein Wert nahe $0$ ist ein Indikator für ein Problem in der Datenqualität. Bewertungskriterien werden mit $q$ bezeichnet und beziehen sich gemäß ihrer Indizierung auf verschiedene Phänomene in den Daten sowie auf eine einzelne Eingangsgröße ($q_{x_j}, j=1,\hdots,p$) oder auf eine bivariate Projektion der Daten auf zwei Eingangsgrößen ($q_{x_j,x_l}, j=1,\hdots,p; l=1,\hdots,p;j\neq l$). Zur Gesamtbewertung einer Eingangsgröße oder zur Gesamtbewertung von Datensätzen mit mehr als 2 Eingangsgrößen können die Bewertungen aller einzelner Eingangsgrößen und aller bivariater Projektionen aggregiert werden.

\subsection{Bewertungskriterien}

\subsubsection{Korrelationen}
Korrelieren Eingangsgrößen des Datensatzes, können die einzelnen Eingangsgrößen univariat gleichverteilt vorliegen, während nur ein kleiner Teil des mehrdimensionalen Eingangsraums mit Daten abgedeckt ist. Eine Korrelation zwischen Eingangsgrößen entspricht einer Redundanz für die Abbildung der Zielgröße, weshalb die Eingangsgrößen für die Modellbildung selektiert oder transformiert und reduziert werden können (PCA-Regression\footnote{PCA = Hauptkomponentenanalyse}, PLS-Regression\footnote{PLS = Partial Least Squares}). Als Hilfsgröße, um Datenqualität bezüglich Korrelationen (engl. \textit{Correlation}) zu quantifizieren, nutzen wir den empirischen Korrelationskoeffizienten $r_{x_j,x_l}$. Daraus berechnet sich das Bewertungskriterium 
\begin{align}
q_{\text{Corr},x_j,x_l} = 1-|r_{x_j,x_l}|.
\end{align}

\subsubsection{Cluster}

Liegen die Daten in Clustern vor, bietet sich das Bilden von lokalen Teilmodellen an. Die Bewertung, ob und wie viele Cluster in einem Datensatz vorliegen, ist ein nichttriviales Problem im Data-Mining-Kontext. In~\cite{Steinbach04} wird die Multimodalität der Häufigkeitsverteilung der paarweisen Distanzen zwischen den Datentupeln als visuelles Kriterium verwendet. Wir quantifizieren die Multimodalität mit Hilfe des \textit{Hartigans DIP Test of Unimodality}~\cite{Hartigan85}. Der DIP Test liefert einen DIP-Index $v_\text{DIP}$ und einen p-Wert $p_\text{DIP}$, die als Indikatoren für Bimodalität, respektive das Vorliegen von Clustern, verwendet werden~\cite{Freeman13}.

\begin{figure}[htb]
\centering
\begin{subfigure}{0.45\textwidth}
\includegraphics[width=\textwidth]{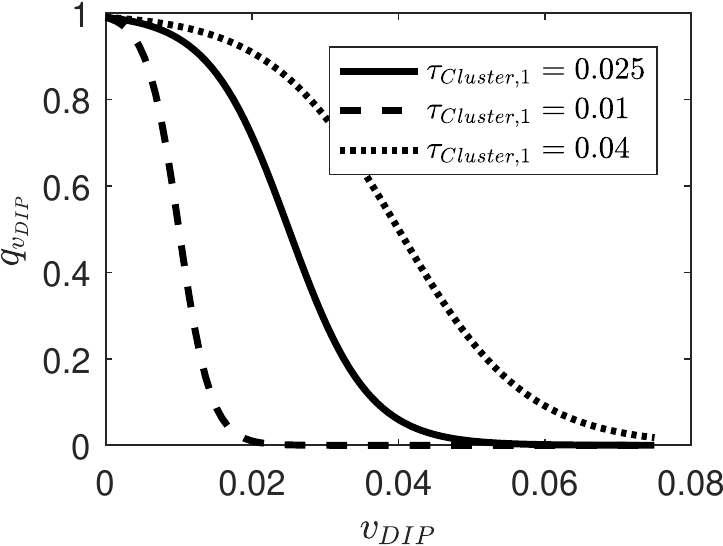}
\subcaption{}
\label{Abb:Guete_Cluster1}
\end{subfigure}
\begin{subfigure}{0.45\textwidth}
\includegraphics[width=\textwidth]{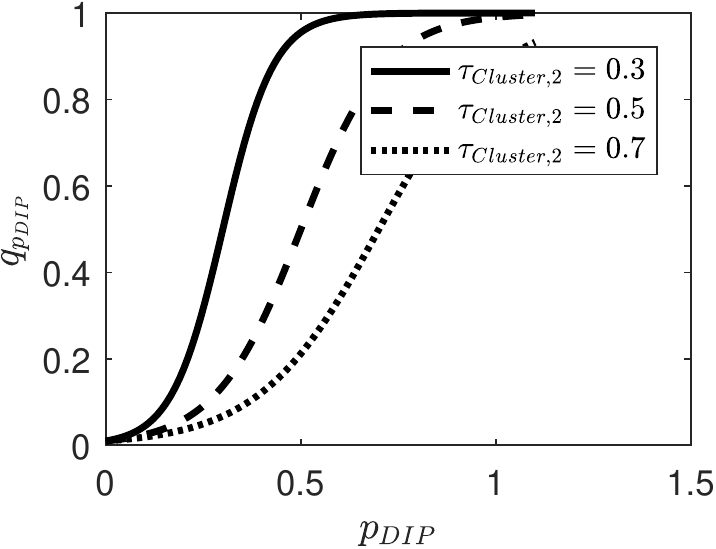}
\subcaption{}
\label{Abb:Guete_Cluster2}
\end{subfigure}
\caption{Verlauf der Sigmoidalfunktionen für das Bewertungskriterium für Cluster}
\label{Abb:Gueten1}
\end{figure}

Wir stellen das Bewertungskriterium
\begin{align}
q_{\text{Cluster},x_j,x_l} = \max(q_{v_\text{DIP},x_j,x_l},q_{p_\text{DIP},x_j,x_l})
\end{align}
mit 
\begin{align}
q_{v_\text{DIP},x_j,x_l} = 1-\frac{1}{1+\exp{\left(-a_1(v_{\text{DIP},x_j,x_l}-\tau_\text{Cluster,1})\right)}}
\end{align}
und
\begin{align}
q_{p_\text{DIP},x_j,x_l} = \frac{1}{1+\exp{\left( -a_2(p_{\text{DIP},x_j,x_l}-\tau_\text{Cluster,2}) \right)}}
\end{align}
vor. $\tau_\text{Cluster,1}$ und $\tau_\text{Cluster,2}$ sind frei parametrierbar. Aus den Randbedingungen $q_{v_\text{DIP},x_j,x_l}(v_{\text{DIP},x_j,x_l}=0)\approx 1$ und $q_{p_\text{DIP},x_j,x_l}(p_{\text{DIP},x_j,x_l}=0)\approx 0$ leiten wir die Parameter
\begin{align}
a_1 = \frac{\ln\left|99\right|}{\tau_\text{Cluster,1}} \hspace{0.5cm}\text{ und } \hspace{0.5cm}a_2 = \frac{\ln\left|99\right|}{\tau_\text{Cluster,2}}
\end{align}
ab. Bild \ref{Abb:Gueten1} zeigt den Verlauf von $q_{v_\text{DIP},x_j,x_l}$ und $q_{p_\text{DIP},x_j,x_l}$ für verschiedene Werte von $\tau_\text{Cluster,1}$ und $\tau_\text{Cluster,2}$. Als Standardwerte werden $\tau_\text{Cluster,1} = 0.025$ und $\tau_\text{Cluster,2} = 0.5$ vorgeschlagen. Die Sigmoidalfunktionen werden verwendet, um das Bewertungskriterium in das interpretierbare Einheitsintervall zu überführen.

\subsubsection{Konfigurationen}

Eingangsgrößen eines Datensatzes können sich darin unterscheiden, wie viele unterschiedliche Ausprägungen von ihnen vorliegen. Durch sehr wenige Ausprägungen einer Eingangsgröße im Verhältnis zu anderen entstehen Cluster. 
 
Sei $c_j$ die Anzahl unterschiedlicher Ausprägungen von $x_j$, dann berechnet sich das Bewertungskriterium
\begin{align}
q_{\text{Config},x_j} = \frac{c_j}{\max_l c_l} \text{, } l=1,\hdots,p.
\end{align}
Ein Wert nahe $0$ kann auf ordinal- oder nominalskalierte Eingangsgrößen hinweisen. Gleichmäßig wenige Ausprägungen aller Eingangsgrößen lassen auf eine statistische Versuchsplanung schließen und sind nicht als Einschränkung in der Datenqualität zu bewerten. Daher wird das Bewertungskriterium in Abhängigkeit zum Maximalwert univariater Ausprägungen ($\max_l c_l$) berechnet.

\subsubsection{Outlier}

\begin{figure}
\centering
\includegraphics[width=0.45\textwidth]{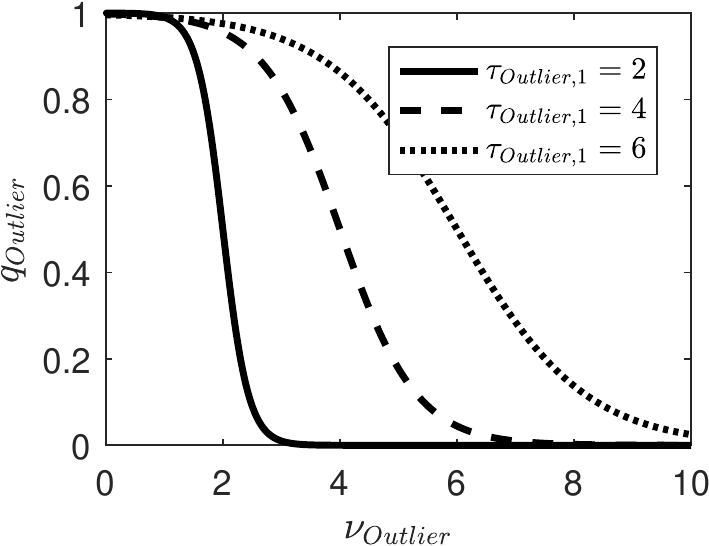}
\caption{Verlauf der Sigmoidalfunktion für das Bewertungskriterium für Ausreißer}
\label{Abb:Guete2}
\end{figure}

Hebelpunkte werden Datentupel genannt, die aufgrund ihrer Lage im Eingangsraum einen großen Einfluss auf die Modellbildung haben. Es handelt sich dabei um sogenannte Ausreißer (engl. \textit{Outlier}). 
Generell werden Ausreißer als Datentupel beschrieben, die sich vom Großteil der anderen Datentupel eines Datensatzes deutlich unterscheiden.
Die Detektion von Ausreißern ist abhängig von der jeweiligen Anwendung. Eine Übersicht über gängige Ansätze findet sich in~\cite{Aggarwal01,Rousseeuw90}.
Bei Ausreißerdetektionen stellt sich anwendungsspezifisch die Frage, ab wann ein Datentupel ein Ausreißer ist, und ob Gruppen von Datentupeln, die entsprechend weit entfernt vom Großteil der Daten liegen, eine Gruppe von Ausreißern darstellt oder bereits ein Datencluster, das nicht von der Modellbildung auszuschließen ist.
Für die Be\-wer\-tung hinsichtlich Ausreißer beinhalte $\boldsymbol{d}^{N\times 1}_{\text{k-NN},x_j,x_l}$ die Distanz jedes Datentupels zu seinem $k$-ten nächsten Nachbarn unter Berücksichtigung der Eingangsgrößen $x_j$ und $x_l$. Der Parameter $k$ bestimmt, wie viele Datentupel eine Gruppe von Ausreißern beinhalten kann, damit sie als solche erkannt wird. Weiterhin sei $d_{\text{k-NN},x_j,x_l,\text{0.9}}$ das $0.9$-Quantil von $\boldsymbol{d}_{\text{k-NN},x_j,x_l}$. Das Quantil lässt sich als maximal zulässiger Anteil der Datentupel verstehen, der als Ausreißer erkannt werden kann. Wir quantifizieren Ausreißer anhand der maximalen Distanz eines Datentupels zu seinem $k$-ten Nachbar $d_{\text{k-NN},x_j,x_l,\text{max}}$: Sei $\nu_{\text{Outlier},x_j,x_l} = \frac{d_{\text{k-NN},x_j,x_l,\text{max}}}{d_{\text{k-NN},x_j,x_l,\text{0.9}}}$, dann wird $\nu_{\text{Outlier},x_j,x_l}$ mit einer Sigmoidalfunktion gemäß 
\begin{align}
q_{\text{Outlier},x_j,x_l} = 1-\frac{1}{1+\exp{-a_3(\nu_{\text{Outlier},x_j,x_l}-\tau_\text{Outlier})}}, \tau_\text{Outlier}>1
\end{align}
in ein Bewertungskriterium überführt. $\tau_\text{Outlier}$ ist frei parametrierbar und bestimmt wie empfindlich die Ausreißerdetektion ist. Aus der Randbedingung $q_{\text{Outlier},x_j,x_l}(\nu_{\text{Outlier},x_j,x_l}=1)\approx 1$ leiten wir den Parameter
\begin{align}
a_3 = -\frac{\ln{99}}{1-\tau_\text{Outlier}}
\end{align}
ab. Bild \ref{Abb:Guete2} zeigt den Verlauf der Gütefunktion für verschiedene Werte von $\tau_\text{Outlier}$. Als Standardwert wird $\tau_\text{Outlier} = 4$ vorgeschlagen. Die Sigmoidalfunktion wird verwendet, um das Bewertungskriterium in das interpretierbare Einheitsintervall zu überführen.

\subsubsection{Orthogonalität}

\begin{figure}[htb]
\centering
\begin{subfigure}{0.3\textwidth}
\includegraphics[width=\textwidth]{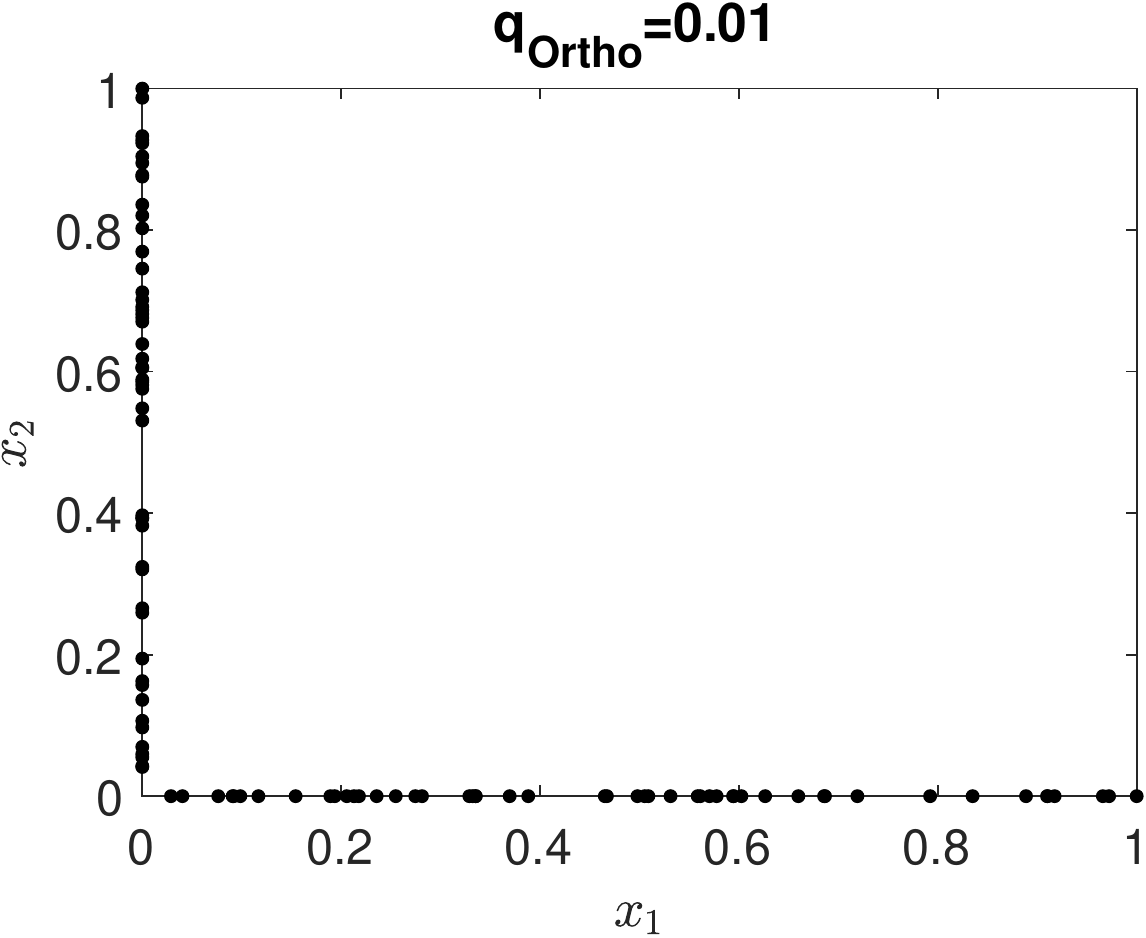}
\subcaption{}
\end{subfigure}
\begin{subfigure}{0.3\textwidth}
\includegraphics[width=\textwidth]{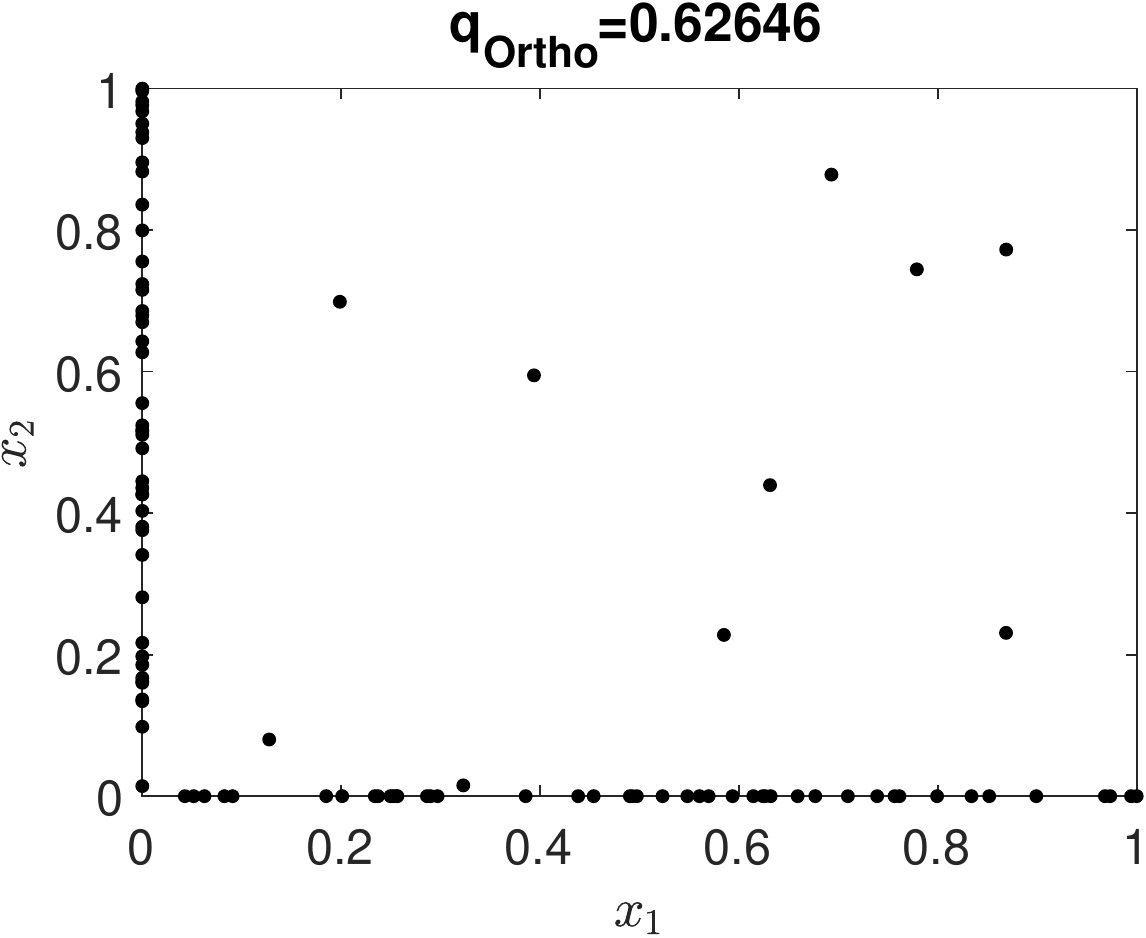}
\subcaption{}
\end{subfigure}
\begin{subfigure}{0.3\textwidth}
\includegraphics[width=\textwidth]{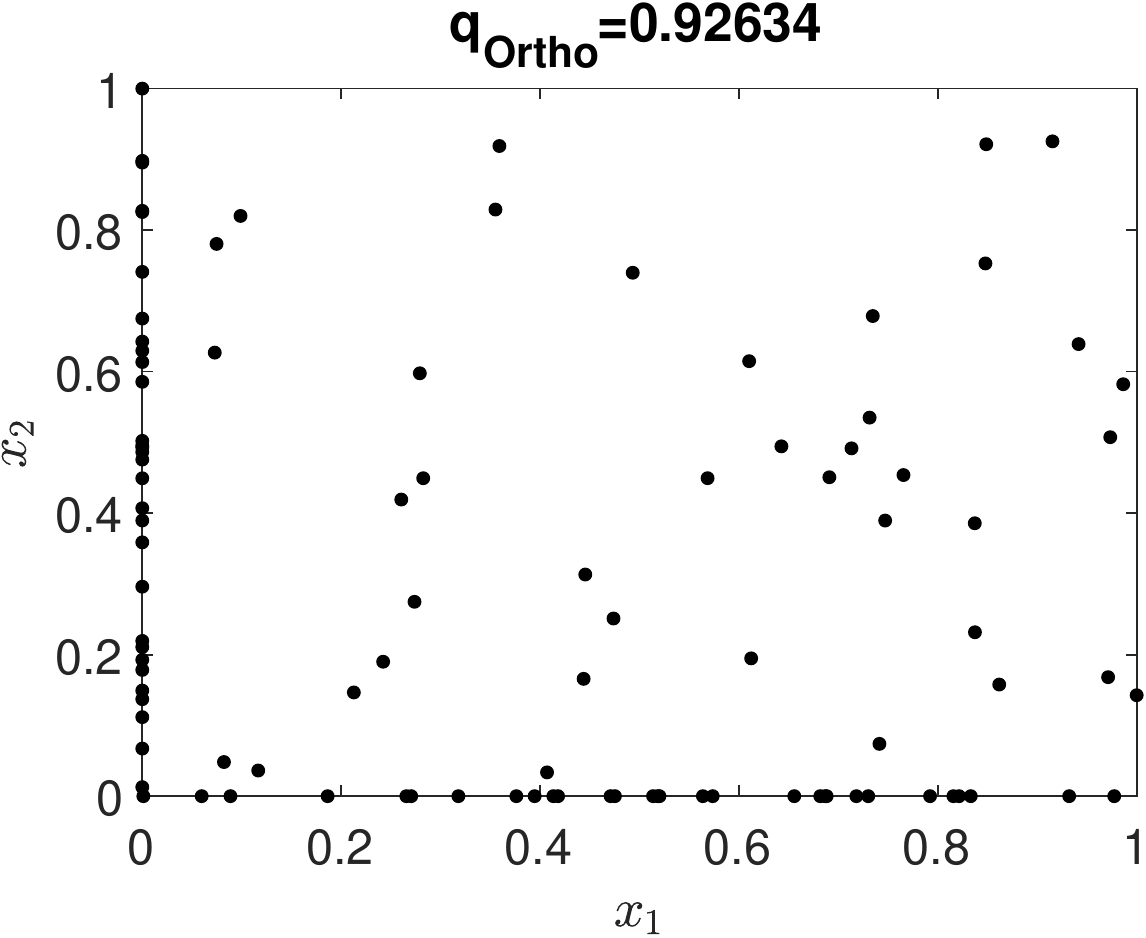}
\subcaption{}
\end{subfigure}
\caption{Streuwolkendiagramme mit abnehmender Orthogonalität}
\label{Abb:Orthogonalitaet_zunehmend}
\end{figure}

Orthogonalität beschreibt das Gegenteil von Korrelationen, wodurch für Regressionen keine Daten vorliegen, die Wechselwirkungen zweier Eingangsgrößen auf die Zielgröße beschreiben. Bild \ref{Abb:Orthogonalitaet_zunehmend} veranschaulicht Streuwolkendiagramme mit verschiedener Ausprägung von Orthogonalität. Bei starker Orthogonalität ist nur ein geringer Teil des zweidimensionalen Eingangsraums mit Daten abgedeckt, obwohl Histogramme beider Eingangsgrößen auf eine ganzheitliche Abdeckung schließen lassen. Da bisher keine Kenngrößen Orthogonalität zuverlässig erkennen können, wird im folgenden Abschnitt ein Bewertungskriterien vorgestellt, um mit einigen Hilfsgrößen ein Maß für Orthogonalität bereitzustellen.

\begin{figure}[htb]
\centering
\includegraphics[width=\textwidth]{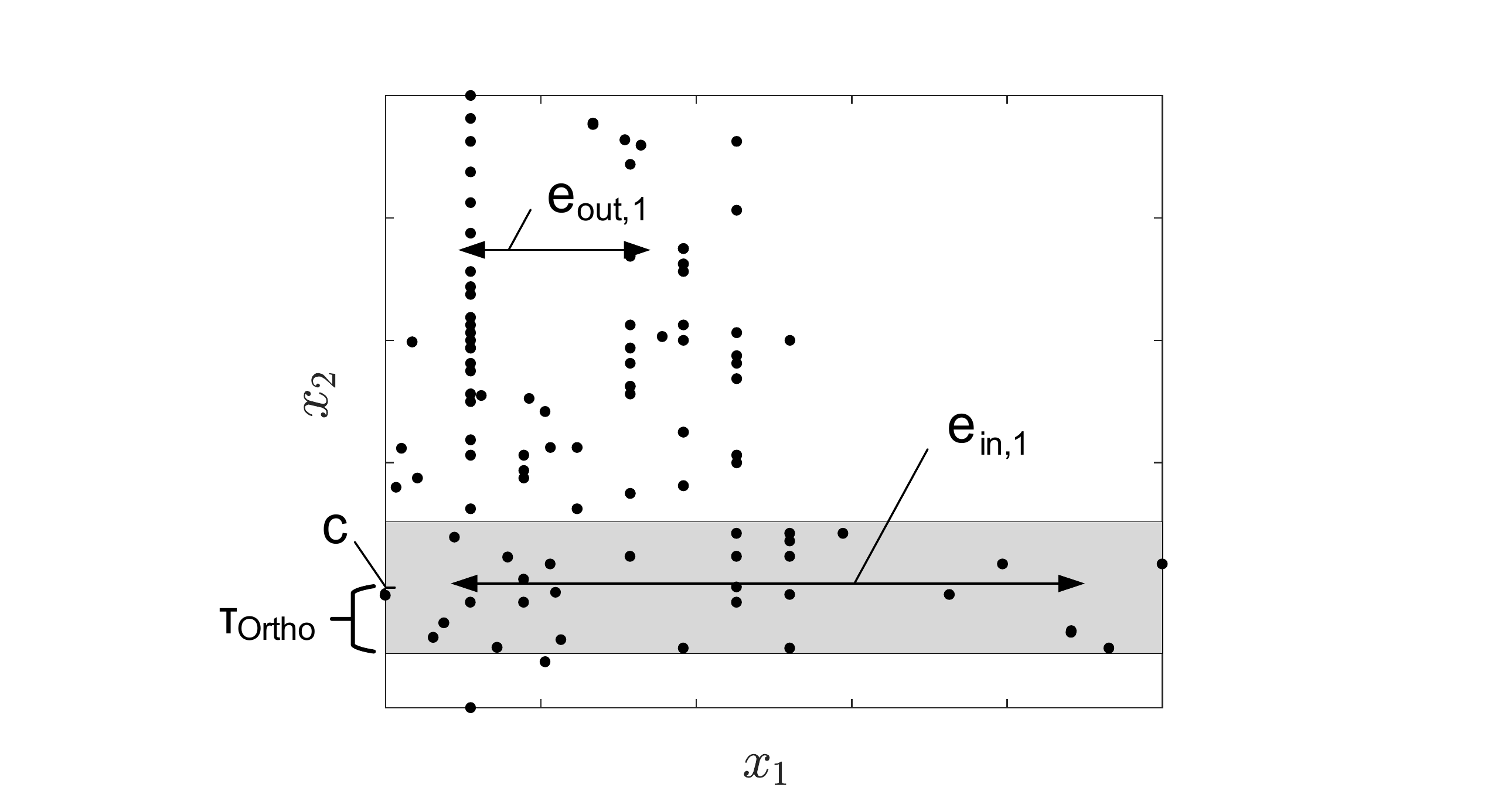}
\caption{Kenngrößen zur Berechnung des Bewertungskriteriums für Orthogonalität}
\label{Abb:Orthogonalitaet_Erklaerung}
\end{figure}

Mit den Indexmengen
\begin{align}
&\mathcal{I}_{\boldsymbol{X}} = \{1,\dots,N\}\text{,} \\
&\mathcal{I}_\text{In} = \{i\in\mathcal{I}_{\boldsymbol{X}}|x_\text{i,l}\in[c-\tau_\text{Ortho},c+\tau_\text{Ortho}]\} \text{ und} \notag\\
&\mathcal{I}_\text{Out} = \mathcal{I}_{\boldsymbol{X}} \setminus \mathcal{I}_\text{In}\notag
\end{align}
ergeben sich die mittleren absoluten Abweichungen
\begin{align}
&e_{\text{Out},j} = \sqrt{\frac{1}{|\mathcal{I}_\text{Out}|}\sum_{i\in\mathcal{I}_\text{Out}}{\left(x_{i,j}-\frac{1}{|\mathcal{I}_\text{Out}|}\sum_{z\in\mathcal{I}_\text{Out}}{x_{z,j}}\right)}} \text{ und}\\
&e_{\text{In},j} = \sqrt{\frac{1}{|\mathcal{I}_\text{In}|}\sum_{i\in\mathcal{I}_\text{In}}{\left(x_{i,j}-\frac{1}{|\mathcal{I}_\text{In}|}\sum_{z\in\mathcal{I}_\text{In}}{x_{z,j}}\right)}} \notag
\end{align}
und das Bewertungskriterium
\begin{align}
q_{\text{Ortho},x_j,x_l} = \min_c{\frac{e_{\text{Out},j}}{e_{\text{In},j}}}.
\end{align}
%
%
%
$\tau_\text{ortho}$ ist ein empirisch zu wählender Parameter, der die Empfindlichkeit des Bewertungskriteriums bestimmt. Für die folgenden Beispiele sei $\tau_\text{ortho}=0.1$. Bild \ref{Abb:Orthogonalitaet_Erklaerung} veranschaulicht die Parameter und Kenngrößen $c,\tau_\text{Ortho},e_{\text{Out},j}$ und $e_{\text{In},j}$.

\section{Beispiele}

\begin{figure}[htb]
\begin{subfigure}{0.3\textwidth}
\includegraphics[width=\textwidth]{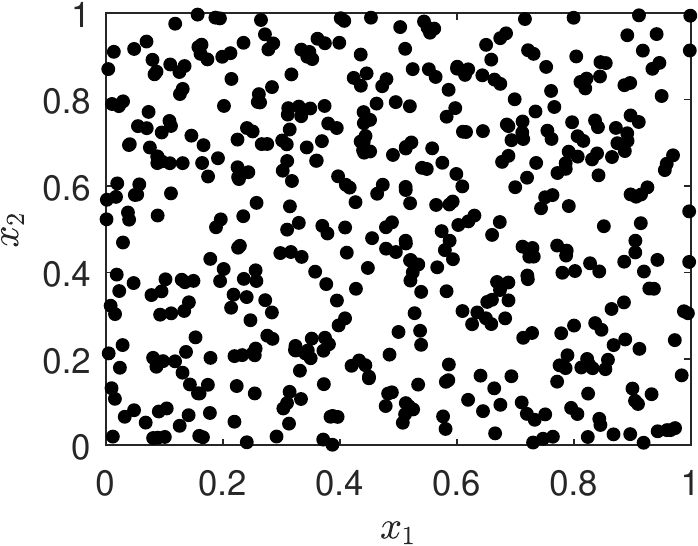}
\subcaption{Vollständigkeit}
\label{Abb:Vollstaendigkeit}
\end{subfigure}
\begin{subfigure}{0.3\textwidth}
\includegraphics[width=\textwidth]{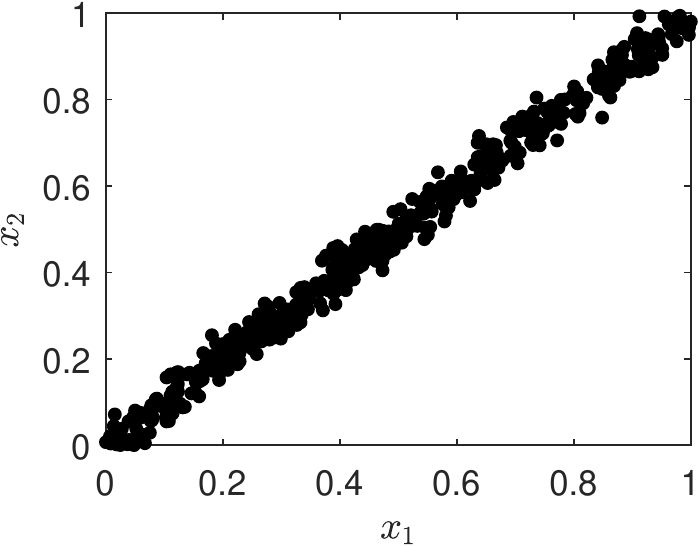}
\subcaption{Korrelation}
\label{Abb:Korrelation}
\end{subfigure}
\begin{subfigure}{0.3\textwidth}
\includegraphics[width=\textwidth]{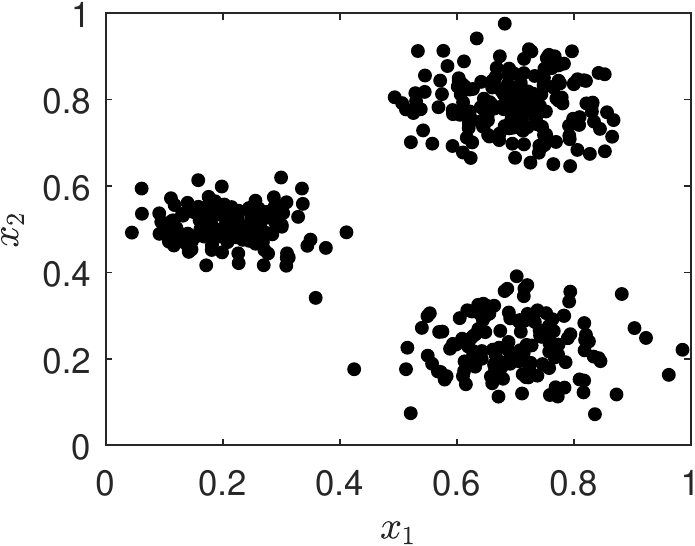}
\subcaption{Cluster}
\label{Abb:Cluster}
\end{subfigure}

\begin{subfigure}{0.3\textwidth}
\includegraphics[width=\textwidth]{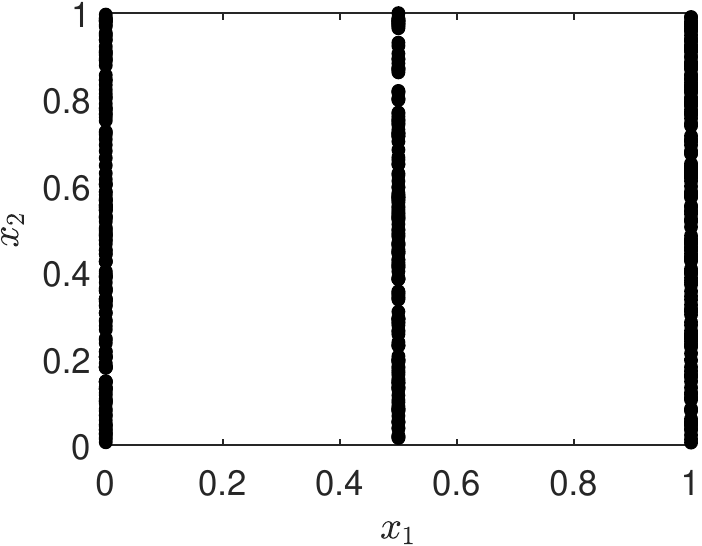}
\subcaption{Konfiguration}
\label{Abb:Konfiguration}
\end{subfigure}
\begin{subfigure}{0.3\textwidth}
\includegraphics[width=\textwidth]{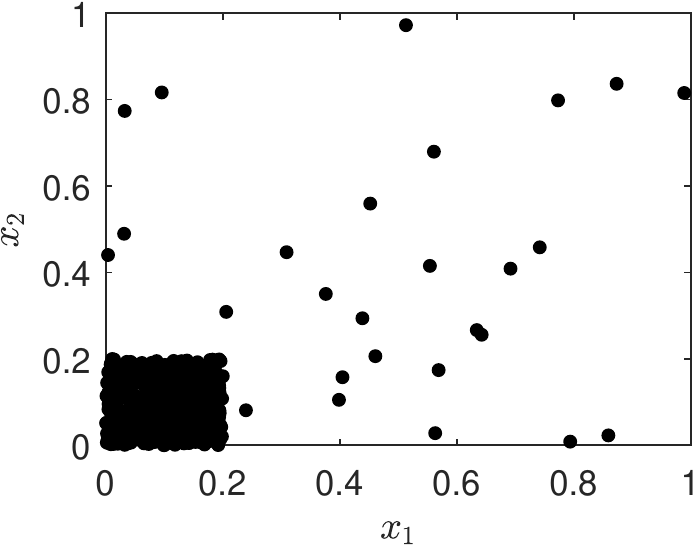}
\subcaption{Ausreißer}
\label{Abb:Ausreisser}
\end{subfigure}
\begin{subfigure}{0.3\textwidth}
\includegraphics[width=\textwidth]{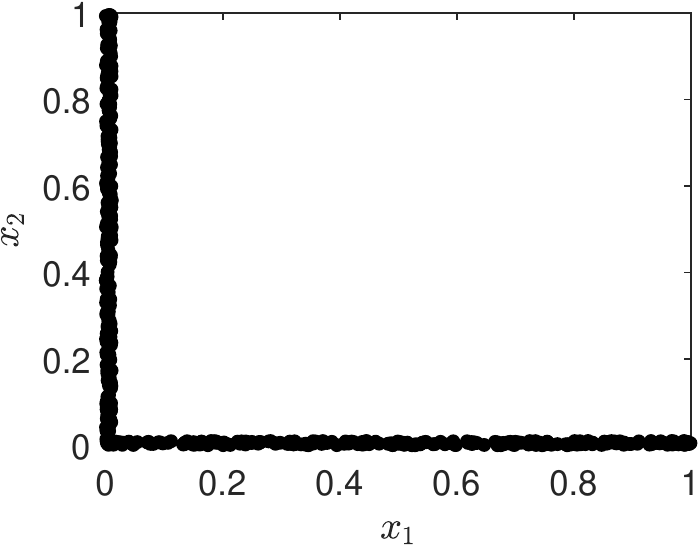}
\subcaption{Orthogonalität}
\label{Abb:Orthogonalitaet}
\end{subfigure}
\caption{Simulierte Benchmark-Datensätze}
\label{Abb:Benchmarks}
\end{figure}

Wir haben Benchmark-Datensätze erstellt, um die Einschränkungen der Datenqualität zu simulieren. Bild \ref{Abb:Benchmarks} zeigt die sechs Benchmark-Datensätze mit jeweils zwei Eingangsgrößen $x_1$ und $x_2$.

Tabelle \ref{Tab:Ergebnisse} zeigt die Bewertungskriterien für die Benchmark-Datensätze. Die unterschiedlichen Phänomene werden getrennt voneinander identifiziert. Lediglich Korrelationen werden auch in zwei anderen Datensätzen detektiert. Das Kriterium $q_\text{Cluster}$ indiziert außerdem Cluster in $\boldsymbol{X}_\text{Configurations}$, was zu erwarten war. Konfigurationen sind auch als Cluster interpretierbar.

\begin{table}[htb]
\centering
\begin{tabular}{l|ccccc}
Datensatz&$q_{\text{Corr},x_1,x_2}$&$Q_{\text{Cluster},x_1,x_2}$&$q_{\text{Config},min}$&$q_{\text{Outlier},x_1,x_2}$&$q_{\text{Ortho},x_1,x_2}$\\\hline
$\boldsymbol{X}_{a}$ & 0.99 & 0.99 & 1.00 & 0.98 & 1.00 \\
$\boldsymbol{X}_{b}$ & 0.00 & 0.99 & 1.00 & 0.94 & 1.00 \\
$\boldsymbol{X}_{c}$ & 0.98 & 0.01 & 1.00 & 0.63 & 1.00 \\
$\boldsymbol{X}_{d}$ & 0.94 & 0.03 & 0.00 & 0.96 & 1.00 \\
$\boldsymbol{X}_{e}$ & 0.49 & 0.98 & 1.00 & 0.00 & 0.73 \\
$\boldsymbol{X}_{f}$ & 0.39 & 0.99 & 1.00 & 0.97 & 0.01 \\
\end{tabular}
\caption{Die vorgestellten Kriterien für simulierte Benchmarkdatensätze. Die Indices der Datensätze beziehen sich auf Bild \ref{Abb:Benchmarks}.}
\label{Tab:Ergebnisse}
\end{table}

\section{Diskussion und Ausblick}
Die Untersuchung der Eingangsdaten ist Bestandteil eines jeden Data-Mining-Prozesses zur Bildung von Klassifikatoren und Regressionen. Die Automatisierung der visuellen Untersuchung der Daten entlastet den Anwender bei Datensätzen und Systemen mit vielen Eingangsgrößen. Die vorgestellten Bewertungskriterien sind in der Lage, in dafür simulierten Benchmark-Datensätzen die unterschiedlichen Phänomene zu erkennen. Eine Implementierung der Bewertungskriterien findet sich in der Open-Source-MATLAB-Toolbox DaMoQ. Eine Beschreibung der Toolbox findet sich in~\cite{Doneit17}. Die Toolbox umfasst außerdem Maße zur Modellvalidierung~\cite{Doneit14}. Geplant sind zudem Erweiterungen, um systematisch Vorwissen in die Modellbildung zu integrieren. Die Integration von Vorwissen stellt eine Möglichkeit dar, schlechte Datenqualität zu kompensieren~\cite{Doneit15Dort,Doneit15b}.

\begingroup
\bibliographystyle{unsrtdin}

\endgroup

\end{document}